# INITIALIZATION OF MULTILAYER FORECASTING ARTIFICIAL NEURAL NETWORKS

**V.V. Bochkarev and Yu.S. Maslennikova**

**Kazan (Volga Region) Federal University**
vbochkarev@mail.ru, YuliaMsl@gmail.com

**Abstract**

In this paper, a new method was developed for initialising artificial neural networks predicting dynamics of time series. Initial weighting coefficients were determined for neurons analogously to the case of a linear prediction filter. Moreover, to improve the accuracy of the initialization method for a multilayer neural network, some variants of decomposition of the transformation matrix corresponding to the linear prediction filter were suggested. The efficiency of the proposed neural network prediction method by forecasting solutions of the Lorentz chaotic system is shown in this paper.

**Keywords:** Artificial neural networks; Forecasting; Neural network initialization; Linear prediction filter

## Introduction

Reliable forecasts are necessary nowadays both in scientific research and daily operations. At present, this problem is usually being solved by the linear prediction method. The latter reduces the solution of a forecasting problem to the search of coefficients of the linear prediction filter of the *p*th order that gives the best prediction of the current value of a real-valued sequence $\hat{x}_n$ from its previous values:

$$\hat{x}_n = a_1 x_{n-1} + a_2 x_{n-2} + \cdots + a_p x_{n-p}; \tag{1}$$

here *p* is the filter order, and the letter *a* denotes coefficients of the filter. As a rule, one minimizes the sum of squared prediction errors considered as a function of the filter coefficients [1]. Since the graph of this function is a paraboloid, the minimum point is unique.

The linear prediction method is efficient for stationary systems, i.e., those systems whose properties do not change in course of time, while the distribution of parameter fluctuations is close to the Gaussian distribution [1, 2]. In more complicated cases one should better use nonlinear methods, e.g., artificial neural networks [3, 4]. However, this method also has some drawbacks. Thus, in the case of neural networks, owing to the nonlinearity of the neurons activation function, the error functional is more complicated, and the calculated local minimum point is not necessary global. As a result, in spite of the great potential of neural networks [5], the error of a neural network prediction model may exceed that of the linear one.

As is mentioned in [3], when using neural networks, the choice of the rule for the initialization of neuron weighting coefficients plays an important role. Algorithms for the initialization of neural networks imply the choice of initial values of weighting coefficients. This choice can be performed by some deterministic technique or random one (e.g., the widely used Nguyen-Widrow algorithm [5]). In the latter case, as a rule, the network is being trained several times in order to increase the probability of finding the global minimum of the error function. This leads to a multiple growth of the computational burden, while the obtained result may appear to be unsatisfactory.

In [6] one proposes an interesting algorithm for constructing a hybrid neural network [4] for time series prediction. First, one trains the network consisting of a single linear neuron. The obtained coefficients are used for initializing a hybrid network, which also contains only one neuron. Then one sequentially adds neurons to the network until the decrease of the prediction error stops. In this paper we consider another approach which is also based on the preliminary solution of the optimal linear prediction problem.

Formula (1) demonstrates that the linear prediction filter is representable as the one-layer linear neural direct transmission network. This analogy allows us to use the preliminarily calculated coefficients of the linear prediction model as a base, including them in the corresponding elements of the neural network. The idea is to construct a neural network which before training transforms data in just the same way as expression (1). In order to make the full use of abilities of neural network algorithms for approximating complex dependencies, the network has to consist of 2 or more layers [3]. And then we meet a difficulty. It is connected with the ambiguity of the decomposition of the initial linear transform into several sequential ones which correspond to layers of the neural network. In what follows we consider two possible ways to overcome this difficulty.

**The simplest algorithm for the initialization of a neural network based on coefficients of the linear prediction model**

Consider the simplest technique for solving the stated problem in the case when functions implemented by various network layers are evidently separated. The error that occurs due to the nonlinearity of neurons activation functions prior to the network training should be small. This fact imposes certain restrictions on the choice of activation functions for layers. For example, the hyperbolic tangent (the sigmoid) has a domain with a weak nonlinearity at the origin of coordinates. When using neurons with sigmoidal activation functions, by properly scaling the training set we can make transforms of a signal by the network neurons close to linear. Let us consider, as an example, the application of the proposed initialization algorithm to a three-layer neural direct transmission network [4].

Assume that input values of all neurons correspond to linear parts of the activation function. Represent the resulting transformation matrix of a three-layer neural network as follows:

$$A = A_3 A_2 A_1, \qquad (2)$$

where $A_1$, $A_2$, $A_3$ are transformation matrices of the 1st, 2nd, and 3d layers, respectively (Fig. 1). The input layer of the network scales the data so as to make all values of the series be located within the linear domain of the hyperbolic tangent.

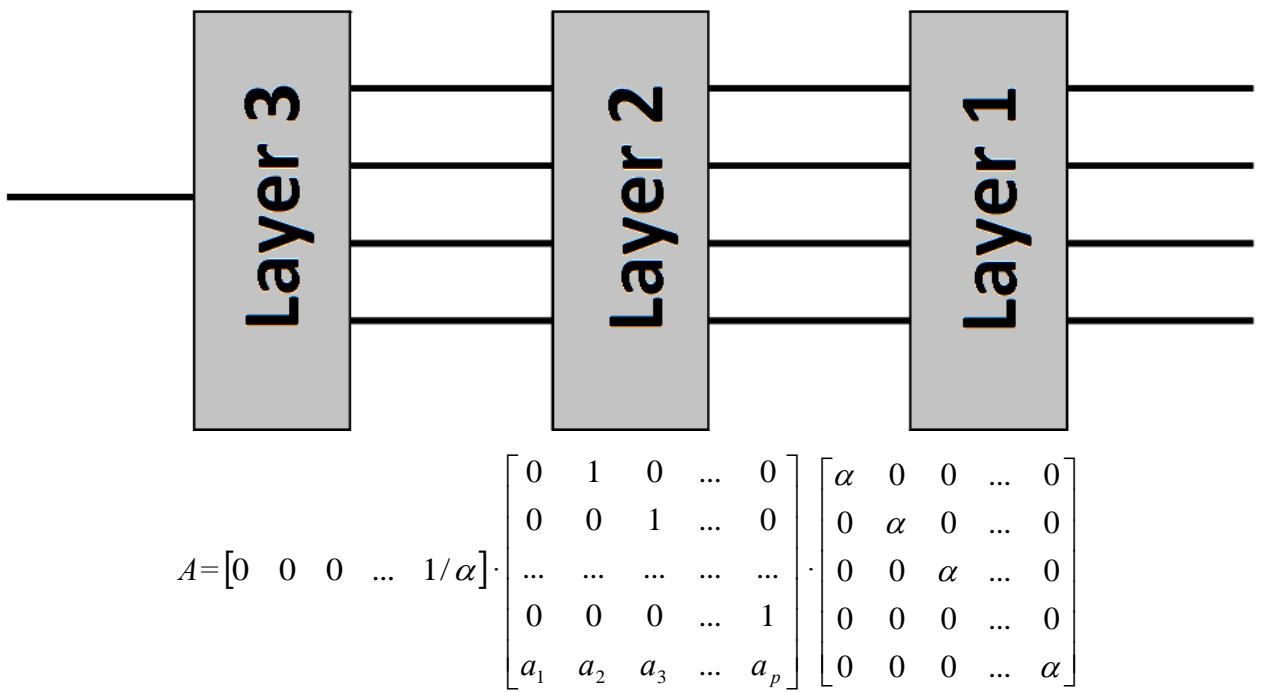

Fig. 1. The transformation matrix of a neural network after the initialization.

The intermediate layer performs a weakly nonlinear transform with the help of the weight matrix that contains coefficients $a$ of the linear prediction model. The output layer performs the converse data transform. By a proper choice of the coefficient $\alpha$ one can make the error stipulated by the nonlinearity of the activation function arbitrarily small. In this case the neural network with weights chosen in such a way will estimate the current value $\hat{x}_k$ of the input signal from preceding $p$ sequential ones $x_{k-1}$, $x_{k-2}$, … $x_{k-p}$ in accordance with expression (1).

Note that results of the prediction of the given neural network before training nearly coincide with the output of the linear prediction filter. When training the network with the help of an algorithm with a monotonic decrease of the error (e.g., gradient and quasi-Newton algorithms, those based on the conjugate gradients method, etc), the result obtained on any training step is the same or even better than that obtained with the help of the linear prediction filter. In this paper we use the Levenberg-Marquardt algorithm; it belongs to the class of quasi-Newton methods that guarantee a high convergence rate [7].

**Prediction of the Lorenz chaotic system**

The proposed technique for the neural network initialization was applied for the prediction of trajectories of the Lorenz system, which is the simplest example of a determinate chaotic system [8]. It was first observed by Lorenz in numerical experiments when studying the trajectory of a system of three connected quadratic ordinary differential equations that define three modes of Oberbeck-Boussinesq equations for the convection of a liquid in a 2D layer heated from below. The mentioned equations take the form

$$\begin{cases} \dot{x} = \sigma(y - x), \\ \dot{y} = x(r - z) - y, \\ \dot{z} = xy - bz. \end{cases} \quad (3)$$

According to results of numerical modeling, solutions of the system with many values of parameters asymptotically tend to unstable cycles, but have two evident clusterization centers and thus form the so-called strange attractor [8]. Fig. 2 illustrates the solution to the system with $\sigma = 10, r = 28, b = 8/3$. Another important property of solutions to the Lorenz system is their essential dependence on the initial condition, which constitutes the main feature of the chaotic dynamics. Therefore, the long-term behavior of the system is poorly predictable.

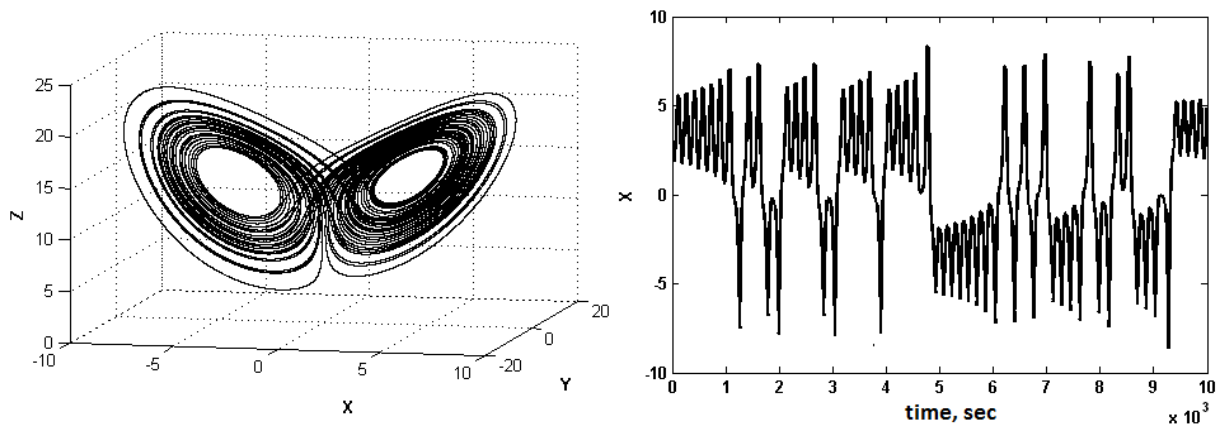

Fig. 2. Trajectories of the Lorenz attractor with r=28 (at the left); the variation of the X coordinate (at the right)

We use the Lorenz system as the test problem due to the fact that it is a rather adequate model for many real systems. In [8, 9] one mentions that this system approximately describes oscillations of parameters in turbulent flows and variations of parameters of the geomagnetic field. One also encounters such system of equations in models of economic processes [10].

For testing the proposed initialization algorithm we have constructed the series that correspond to solutions of system (3). We have applied the Runge–Kutta method with the automatic choice of the step value. We have thinned out the obtained series in order to get a sample of an acceptable size covering a long time interval with the time step of 0.01 sec. Moreover, we have removed the initial part of the sample in order to make the data used for testing correspond to the movement near the attractor and to make the results independent of the choice of the initial point.

Below we represent results of the prediction of the behavior of the given system with the help of the neural network, where initial values of weights are chosen in accordance with coefficients of the linear prediction filter. First we have calculated coefficients of the linear prediction model and then on its base constructed the neural network in the way described above. We predicted each current value from 5 sequential previous ones. The first and the second layer of the neural network contained 5 neurons each. We trained the neural network with the help of the error backpropagation method, minimizing the error functional by the Levenberg-Marquardt algorithm within 300 epochs of training. The prediction was performed for a time interval varying from 0.01 to 1 second.

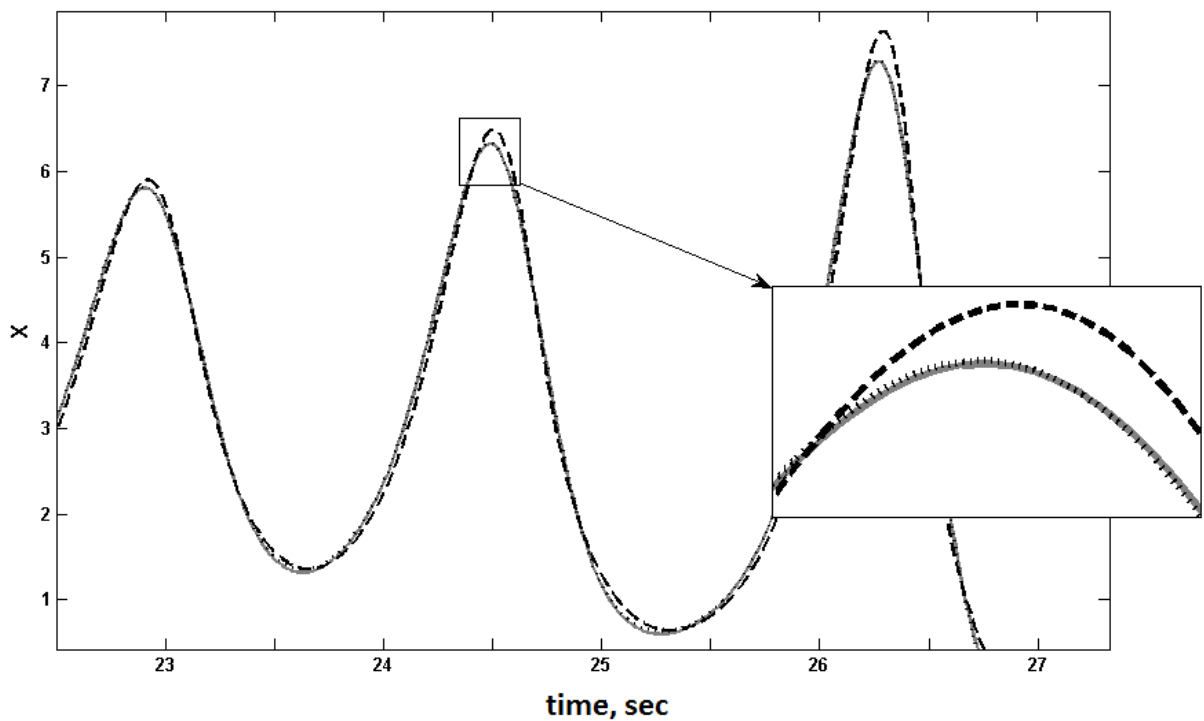

Fig. 3. A part of the predicted series for the increase of the X coordinate by 0.15 seconds; the interior of the rectangle shows a zoomed part of the same series: the initial series (solid line), the linear prediction (dashed line), the neural network prediction (dotted line)

Fig. 3 shows a part of the predicted series; for comparison we also depict the linear prediction of the same series. This example is typical for the whole range of the prediction interval, namely, the proposed initialization algorithm provides an essential improvement in comparison both to the linear prediction method and to the neural network prediction method used in the Nguyen-Widrow initialization algorithm. See Table 1 below for mean square prediction errors for various time intervals in comparison to results obtained by other methods.

The results of numerical experiments prove the efficiency of the proposed neural network initialization method. We can also ascertain that for the considered problem the neural network prediction is much more efficient than the linear prediction method.

### An improved network initialization method

The main drawback of the considered simplest algorithm consists in the fact that layers of the network (at least, on the initial training stage) participate to some extent in the data transformation. In order to fully use abilities of multilayer networks, it is desirable to make the computational load of layers uniform. As was mentioned above, the described method for the initialization of a neural network is not unique. According to the considered example of a three-layer neural network, one can include real orthogonal matrices $U_1$ and $U_2$ in the resulting transformation matrix $A$ (see expression (2) and Fig. 1). Then $U_1^T U_1 = E$, $U_1^T U_1 = E$, and

$$A = A_3 U_2^T U_2 A_2 U_1^T U_1 A_1. \tag{4}$$

Therefore, before training, the neural network, where initial values of weighting coefficients of layers are given by matrices $U_1 A_1$, $U_2 A_2 U_1^T$, and $A_3 U_2^T$, respectively, still completely corresponds to the initial linear prediction filter. However, during the training process, because of the nonlinearity of the neurons transfer function, we can expect an improvement of the prediction accuracy. In order to choose optimal matrices $U_1$ and $U_2$, it is convenient to represent them in the exponential form

$$U_i = e^{G_i}. \tag{5}$$

As is known, in this case, in order to make matrices $U_i$ orthogonal, matrices $G_i$ have to be skew-symmetric

$$G_i^T = -G_i.$$

Let the number of neurons in the 1st and 2nd layers equal $N$. In the linear space of skew-symmetric $N \times N$-matrices (the dimension of this space equals $N(N-1)/2$) we choose some basis, for example,

$$G^{(k,m)}, k < m, \quad G_{i,j}^{(k,m)} = \delta_{m,i} \delta_{k,j} - \delta_{k,i} \delta_{m,j}.$$

Then
$$G_i = \sum_{k,m} x_{k,m,i} G^{(k,m)}. \tag{6}$$

For example, with $N = 3$ we have

$$G_i = \begin{pmatrix} 0 & a & b \\ -a & 0 & c \\ -b & -c & 0 \end{pmatrix} = a_i \begin{pmatrix} 0 & 1 & 0 \\ -1 & 0 & 0 \\ 0 & 0 & 0 \end{pmatrix} + b_i \begin{pmatrix} 0 & 0 & 1 \\ 0 & 0 & 0 \\ -1 & 0 & 0 \end{pmatrix} + c_i \begin{pmatrix} 0 & 0 & 0 \\ 0 & 0 & 1 \\ 0 & -1 & 0 \end{pmatrix}.$$

If the number of neurons in layers is small, then one can determine unknowns in expression (4) by the gradient descent method [7]. The most appropriate choice in this case is to define the objective function as the error of the neural network which is initialized in the mentioned way and then subject to the full course of training. However, it is clear that this approach requires more calculations. In this case, only the calculation of the gradient from coefficients $x_{k,m,i}$ that enter in (6) requires $\sim N^2$ network training courses.

One can solve this problem either by using stochastic search algorithms or by the random initialization of coefficients $x_{k,m,i}$. In this case instead of the exponential parameterization of orthogonal matrices (5) one can perform the parameterization

$$U_i = (I - G_i)(I + G_i)^{-1},$$

which requires less computations.

The search of the global minimum in this case requires multiple training of the neural network. Another approach (we describe it below) is based on the following ideas. Let us choose matrices $U_1$ and $U_2$ so as to make the nonlinear transformation of data in the neural network bring the gain as soon as possible. The rate of the neural network training depends on the gradient calculated by differentiating the error function with respect to neurons weights. We vary coefficients $x_{k,m,i}$ in (6) so as to maximize the norm of the gradient. Note that this process is quite analogous to the training of the neural network by the error backpropagation method, because, in essence, we just impose some linear constraints on the variation of weighting coefficients in accordance with expressions (4). In view of this fact we expect the total time consumption to be approximately two times greater than that of the simplest initialization algorithm described above. The implementation of this algorithm needs minimal additions to available libraries of neural network computations. For example, the network error function after the first epoch of training can be used as the objective function. In both cases, one can solve the optimization problem, for example, by the gradient descent method.

We have performed numerical experiments with the proposed algorithm for the same model data and the same structure of the neural network as in the previous case. For simplicity of programming, we used the error of the neural network after the first epoch of training as the objective function. We performed 10 iterations of the gradient descent method. Then for the network with calculated optimal values of parameters of initial weights we, as above, performed 300 epochs of training. With fixed parameters the total time consumption increased less than 2 times in comparison to the simplest initialization algorithm considered above.

See Table 1 for values of mean square prediction errors obtained by the application of 4 methods (the linear prediction method, the neural network prediction method with the Nguyen-Widrow initialization algorithm, the simplest algorithm, and the improved one based on the use of coefficients of the linear prediction model) under the same conditions for prediction intervals of 0.01, 0.02, 0.05, 0.1, and 1 second. Evidently, the use of the improved initialization algorithm in all cases results in the further essential decrease of the prediction error.

Table 1

| **Prediction method** | **σ (0.01 sec)** | **σ (0.02sec)** | **σ (0.05 sec)** | **σ (0.1 sec)** | **σ (1 sec)** |
|---|---|---|---|---|---|
| **Linear prediction method** | 3.138·10⁻⁴ | 13·10⁻⁴ | 0.0108 | 0.0646 | 3.4031 |
| **Neural network prediction method** | 1.184·10⁻⁴ | 9.441·10⁻⁴ | 0.0062 | 0.0377 | 0.7285 |
| **Neural network prediction method based on the linear one** | 0.505·10⁻⁴ | 1.997·10⁻⁴ | 0.0016 | 0.0097 | 0.0862 |
| **Improved neural network prediction method** | 0.275·10⁻⁴ | 0,985·10⁻⁴ | 0.0009 | 0.0048 | 0.0564 |

For comparison we have also determined values of coefficients $x_{k,m,i}$ by minimizing the error at the output of the fully trained network. In this case the error of the prediction for 0.05 sec equals 0.00072 against the value of 0.0009 obtained by using the greatest gradient criterion. Evidently, switching to the method that requires much more computations, we only insignificantly improve the prediction accuracy.

## Conclusion

According to the predicted behavior of the Lorenz chaotic system, the proposed initialization method for a neural network allows one to essentially decrease the prediction error in comparison both to that in the linear prediction method and the neural network one under the use of universal initialization algorithms. It is also true that in this problem, under the same conditions, the neural network prediction method is much more efficient than the linear one. The improved initialization algorithm allows us to additionally improve the prediction accuracy by a slight increase of the training time. Therefore, the developed method proved to be consistent for predicting the behavior of trajectories of the Lorenz chaotic system, which allows one to use it for predicting the behavior of such complex series.